\documentclass[10pt, a4paper]{article}

\usepackage[]{lrec-coling2024} %

\usepackage{graphicx}
\usepackage{picinpar}
\usepackage{subcaption}
\usepackage{colortbl}
\usepackage{booktabs}
\usepackage{multirow}
\usepackage{soul}
\usepackage{enumerate}
\usepackage{color}
\usepackage{hyperref}
\usepackage{amsmath}
\usepackage{amssymb}
\usepackage{algorithm}
\usepackage{algpseudocode}
\usepackage{mathabx}

\title{Knowledge-augmented Graph Neural Networks with Concept-aware Attention for Adverse Drug Event Detection} %

\name{Yao Gao~\textsuperscript{1~$\asterisk$}, Shaoxiong Ji~\textsuperscript{2~$\asterisk$}\thanks{$\asterisk$~Equal contribution. Corresponding author: Shaoxiong Ji. Work done while affiliated with Aalto University.}, Pekka Marttinen~\textsuperscript{1}} 

\address{\textsuperscript{1} Aalto University \quad \textsuperscript{2} University of Helsinki \\
     \{yao.gao;~pekka.marttinen\}@aalto.fi,~shaoxiong.ji@helsinki.fi\\}

\abstract{
Adverse drug events (ADEs) are an important aspect of drug safety. 
Various texts such as biomedical literature, drug reviews, and user posts on social media and medical forums contain a wealth of information about ADEs. 
Recent studies have applied word embedding and deep learning-based natural language processing to automate ADE detection from text. 
However, they did not explore incorporating explicit medical knowledge about drugs and adverse reactions or the corresponding feature learning. 
This paper adopts the heterogeneous text graph, which describes relationships between documents, words, and concepts, augments it with medical knowledge from the Unified Medical Language System, and proposes a concept-aware attention mechanism that learns features differently for the different types of nodes in the graph. 
We further utilize contextualized embeddings from pretrained language models and convolutional graph neural networks for effective feature representation and relational learning.
Experiments on four public datasets show that our model performs competitively to the recent advances, and the concept-aware attention consistently outperforms other attention mechanisms. 
 \\ \newline  }

\begin{document}

\maketitleabstract

\section{Introduction}

Pharmacovigilance, i.e., drug safety monitoring, is a critical step in drug development~\citep{wise2009new}. 
It detects adverse events and safety issues and promotes drug safety through post-market assessment; therefore, it promotes safe drug development and shows significant promise in better healthcare service delivery.
A drug-related negative health outcome is referred to as an Adverse Drug Event (ADE)~\citep{donaldson2000err}. 
Given the significant harm caused by ADEs, it is essential to detect them for pharmacovigilance purposes. 

Clinical trials are the common way to detect ADEs. 
However, some ADEs are hard to investigate through clinical trials due to their long latency~\citep{sultana2013clinical}. 
Additionally, regular trials cannot cover all aspects of drug use. 
Through the voluntary Post-marketing Drug Safety Surveillance System~\citep{li2014adverse}, users report their experiences with drug usage and related safety issues. 
Nevertheless, the system suffers several limitations, such as incomplete reporting, under-reporting, and delayed reporting.  

\begin{figure}[ht]
\centering
\includegraphics[width=0.5\textwidth]{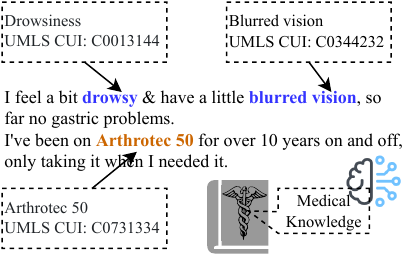}
\caption{An example of a text mentioning an adverse drug event from \citet{karimi2015cadec}. The recognition of drugs and adverse reactions requires medical knowledge and relational reasoning.}
\label{fig:example}
\end{figure}

Recent advances in automated pharmacovigilance are based on collecting large amounts of text about adverse drug events from various platforms, such as medical forums (e.g., AskaPatient), biomedical publications, and social media, and training Natural Language Processing (NLP) models to automatically detect whether a given textual record contains information about adverse drug reactions, which is usually framed as a binary classification task.
Text mentions of adverse drug events include a plethora of drug names and adverse reactions.
Figure~\ref{fig:example} shows an example annotated with concepts from the Unified Medical Language System (UMLS), where each identified medical concept in the text is assigned a Concept Unique Identifier (CUI) from UMLS.
To understand the drug information and corresponding adverse reactions, the NLP model needs to capture abundant medical knowledge and be able to do relational reasoning.  

Early studies used rule-based methods \citep{xu2010medex,sohn2014medxn} with manually built rules or applied machine learning algorithms such as conditional random fields~\citep{nikfarjam2015pharmacovigilance,wang2022explainable}, support vector machine~\citep{bollegala2018learning}, and neural networks~\citep{cocos2017deep,huynh2016adverse}. 
These approaches can process text with manual feature engineering or enable automated feature learning with deep learning methods, allowing for automated ADE detection. However, they are limited in capturing rich contextual information and relational reasoning.

Graphs are expressive and can represent various data. 
For example, nodes in a graph for a collection of texts can represent various entities, such as words, phrases, and documents, while edges represent relationships between them.
Such text graphs, together with graph neural networks, are widely used in NLP applications such as sentiment classification and review rating \citep{yao2019graph,lin2021bertgcn,zhang2020every}. 
Recently, graphs have been used for text representation with graph boosting~\citep{shen2020graph} or contextualized graph embeddings~\citep{gao2022contextualized} for ADE detection.
Other works have applied knowledge graph embeddings and link prediction to ADE prediction in drug-effect knowledge graphs~\citep{kwak2020drug,joshi2022knowledge}.
However, medical knowledge plays an important role in ADE detection from text, and so far, there are no studies that incorporate medical knowledge in a text graph and learn concept-aware representations that inject medical concepts (e.g., the UMLS concepts as illustrated in Figure~\ref{fig:example}) into the text embeddings.

A recent model called CGEM~\citep{gao2022contextualized}, which is close to our work, applied a heterogeneous text graph, embodying word and document relations for an ADE corpus, to learn contextualized graph embeddings for ADE detection. Here, we extend the work by \citet{gao2022contextualized} in two useful ways. First, we show how the graph can be augmented with medical knowledge from the UMLS metathesaurus~\citep{umls} and how different similarity measurement methods can contribute to better knowledge infusion. Second, we deploy concept-aware self-attention that applies different feature learning for various types of nodes.
We name our model as KnowCAGE (Knowledge-augmented Concept-Aware Graph Embeddings).
Our contributions are thus summarized as follows: 
\begin{itemize}
    \item We introduce medical knowledge, i.e., the UMLS metathesaurus, to augment the contextualized graph embedding model for representation learning on drug adverse events.
    \item A concept-aware self-attention is devised to learn discriminable features for the concept (from the medical knowledge), word, and document nodes. 
    \item Experimental results evaluated in four public datasets from medical forums, biomedical publications, and social media show our approach outperforms recent advanced ADE detection models in most cases. 
\end{itemize}

\section{Related Work}
\label{sec::related}
Recent advances in adverse drug event detection use word embeddings and neural network models to extract text features and capture the drug-effect interaction. 
Many studies deploy recurrent neural networks to capture the sequential dependency in text.
For example, \citet{cocos2017deep} utilized a Bidirectional Long Short-Term Memory (BiLSTM) network, and \citet{luo2017recurrent} proposed to learn sentence- and segment-level representations based on LSTM.
To process entity mentions and relations for ADE detection and extraction, pipeline-based systems \citep{dandala2019adverse} and jointly learning methods \citep{wei2020study} are two typical approaches. 

Several recent publications studied graph neural networks for ADE detection.
\citet{kwak2020drug} built a drug-disease graph to represent clinical data for adverse drug reaction detection.
GAR~\citep{shen2021gar} uses graph embedding-based methods and adversarial learning. 
CGEM~\citep{gao2022contextualized} combines contextualized embeddings from pretrained language models with graph convolutional neural networks.

Some other studies also adopted other neural network architectures, such as capsule networks and self-attention mechanisms.
\citet{zhang2020gated} proposed the gated iterative capsule network (GICN) using CNN and a capsule network to extract the complete phrase information and deep semantic information. 
The gated iterative unit in the capsule network enables the clustering of features and captures contextual information.
The attention mechanism prioritizes representation learning for the critical parts of a document by assigning them higher weight scores. 
\citet{ge2019detecting} employed multi-head self-attention, and \citet{wunnava2020dual} developed a dual-attention mechanism with BiLSTM to capture semantic information in the sentence. 

Another direction of related work is knowledge augmentation for deep learning models. 
Many publications adopt knowledge graph to guide the representation learning in various applications~\citep{ji2022survey}. 
For example, \citet{ma2018targeted} injected commonsense knowledge into a long short-term memory network, and \citet{liang2022aspect} enhanced the graph convolutional network with affective knowledge to improve aspect-based sentiment analysis. 
Knowledge injection is also used for other applications such as hate speech detection~\citep{pamungkas2021joint}, mental healthcare~\citep{yang2022mental}, and personality detection~\cite{poria2013common}.
Some other works focus on the construction of medical knowledge graphs and data augmentation with knowledge graphs~\cite{sun2020medical,shi2022mda}.

\section{Methods}
\label{sec::methods}

\begin{figure*}[ht!]
    \centering
    \includegraphics[width=\textwidth]{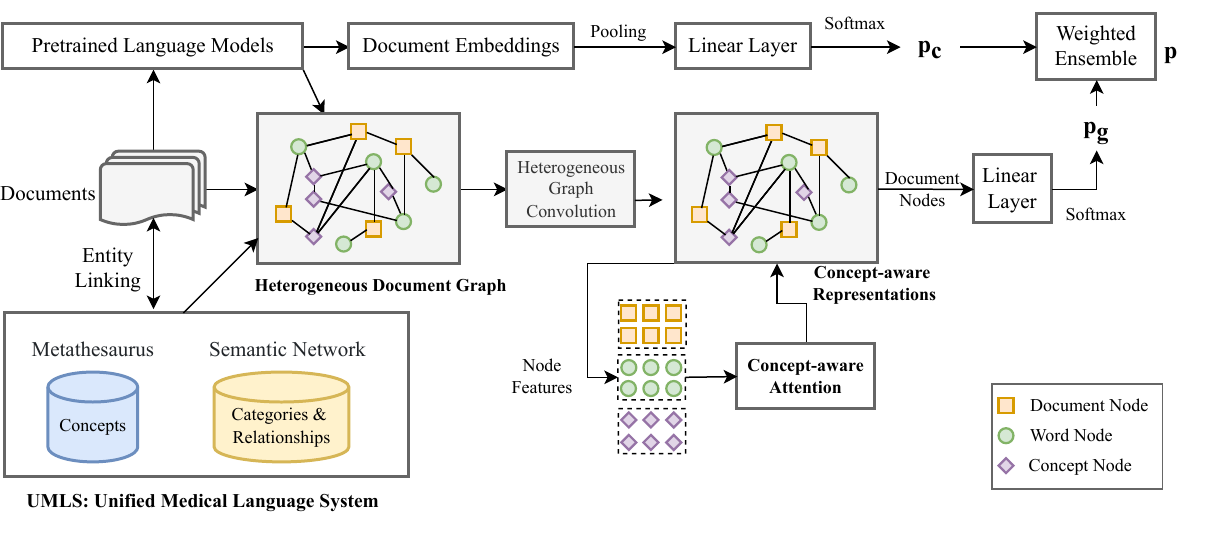}
    \caption{An illustration of the model architecture with knowledge-augmented graph embeddings and concept-aware representations}
    \label{fig:model_KnowCAGE}
\end{figure*}

This section introduces the proposed graph-based model with knowledge augmentation, i.e., Knowledge-augmented Concept-Aware Graph Embeddings (KnowCAGE), as illustrated in Figure~\ref{fig:model_KnowCAGE}.
The model consists of four components: 1) Knowledge-augmented Graph Construction, 2) Heterogeneous Graph Convolution, 3) Concept-aware Attention, and 4) Ensemble-based ADE classification layers.
Following TextGCN~\citep{yao2019graph}, we construct a heterogeneous text graph, which contains three types of nodes: words, documents, and concepts, and we augment it with medical knowledge from the Unified Medical Language System metathesaurus. 
Heterogeneous graph convolution networks are then used to encode the text graph and learn rich representations.
We use the contextualized embeddings from pretrained BERT (Bidirectional Encoder Representations from Transformers)~\citep{devlin2018bert}  to represent the node features in the heterogenous text graph.
The adjacency matrix and feature matrix obtained from the embedding layers are inputs to graph neural network encoders, which take into account the relationships and information between and within the nodes. 
Considering different types of nodes, we use a concept-aware self-attention, inspired by the entity-aware representation learning~\citep{yamada2020luke}, which treats the different types of nodes differently, allowing the most significant content to have the largest contribution to the final prediction.
To boost the prediction of ADE even further, we follow the BertGCN model \citep{lin2021bertgcn} and apply an ensemble classifier with contextualized embeddings on the one hand and the graph networks on the other, and learn a weight coefficient to balance these two prediction branches.

\subsection{Knowledge-augmented Graph Construction}
We firstly build the heterogeneous text graph for the whole document collection by using the external knowledge source - UMLS - to augment the word/document graph with concept information. 
Representing text in a heterogeneous graph can provide different perspectives for text encoding and improve ADE detection.
In the UMLS metathesaurus, different words or phrases are assigned CUIs, where each CUI represents one concept class. 
Every concept class has an attribute ``preferred name'' which is a short description or a synonym of this concept. 
Given a document, our initial step involves employing the longest string match to map words or phrases in the document to CUIs.
Within the UMLS, numerous concepts may be associated with a single word or phrase. We utilize the first concept returned by the system.
Note that we also filter out concepts falling under certain semantic types, such as those related to plants and animals, which are irrelevant to the task. 
As a result of this mapping strategy, only a small subset of concepts are extracted for each dataset.

After our model leverages UMLS to retrieve the preferred names corresponding to concepts in the dataset, these preferred names are added to the graph as the concept nodes. 
Therefore, the augmented graph also contains concept nodes in addition to the word and document nodes. 
The number of total nodes $n=n_{d}+n_w+n_c$, where  $n_d$, $n_w$ and $n_c$ are the numbers of documents, words, and concepts, respectively. 
There are five types of edges, i.e., word-word, word-concept, document-concept, concept-concept, and document-word edges. 
The weights of document-word edges and document-concept edges are calculated as the term frequency-inverse document frequency (TF-IDF), while the weights of the other edges are defined as the similarity or association between two nodes.

Depending on the characteristics of the data set, the most suitable way of knowledge infusion varies.
Therefore, we explore different measurement methods (represented as $\operatorname{SIM}$): L1 distance, L2 distance, Cosine distance, and Pointwise Mutual Information.
They provide distinct insights about node relations, such as co-occurrence information or knowledge about semantic distances.

The word encoding for distance computation is derived from fine-tuning language model on ADE classification. 
Specifically, the weight between the node $i$ and the node $j$ is computed as:
\begin{equation*}
\small
\mathbf{A}_{ij}=
\left\{  
             \begin{array}{lr}  
             \operatorname{SIM}(i,j), & \operatorname{SIM}>0; \text{i,~j:~word/concept}\\  
             \operatorname{TF-IDF_{ij}}, & \text{i:~document,~j:~word/concept}\\ 
             0, & \text{otherwise}
             \end{array}  
\right.
\end{equation*}

We use pretrained contextualized embeddings from language models. 
Given the dimension of embeddings denoted as $d$, the pooled output of contextualized document encoding is denoted as $\mathbf{H}_{doc} \in \mathbb{R}^{{n_d}\times d} $. 
We initialize word and concept nodes with a zero matrix to get the initial feature matrix which is used as input to the graph neural network:
\begin{equation}
\mathbf{H}^{[0]}=
\left (\begin{array}{cc}
 \mathbf{H}_{doc} \\
\mathbf{0}  \\
\end{array}\right),
\end{equation}
where $\mathbf{H}^{[0]} \in \mathbb{R}^{{(n_d+n_w+n_c)}\times d}$ and $[0]$ denotes the initial layer.
\subsection{Heterogeneous Graph Convolution}
We adopt graph neural networks over the heterogeneous text graph to learn complex relations between words, concepts, and documents. 
Specifically, given the initial input features $\mathbf{H}^{[0]}$  obtained from pretrained language models and the adjacency matrix $\mathbf{A}$, we update the representations via graph convolution. 
A forward pass of the $i$-th layer of a convolutional graph network can be denoted as:
\begin{equation}
\label{eq:gnn}
    \mathbf{H}^{[i+1]}=f\left(\mathbf{\hat{A}} \mathbf{H}^{[i]} \mathbf{W}^{[i]} \right),
\end{equation}
where $\mathbf{\hat{A}}$ is the normalized adjacency matrix, $\mathbf{H}^{[i]}$ are the hidden representations of $i$-th layer, $\mathbf{W}^{[i]}$ is the weight matrix, and $f(\cdot)$ is an activation function.  
The KnowCAGE framework can adopt various types of convolutional graph neural networks.
Our experimental study chooses three representative models, i.e., Graph Convolutional Network (GCN)~\citep{kipf2016semi}, Graph Attention Network (GAT)~\citep{velickovic2018graph}, and Deep Graph Convolutional Neural Network (DGCNN)~\citep{zhang2018end}. 
GCN is a spectral-based model with a fixed number of layers where different weights are assigned to layers and the update of node features incorporates information from the node's neighbors.
It employs convolutional architectures to get a localized first-order representation.
Graph attention layers in GAT assign different attention scores to one node's distant neighbors and prioritize the importance of different types of nodes.
DGCNN concatenates hidden representations of each layer to capture rich substructure information and adopt a SortPooling layer to sort the node features. 

\subsection{Concept-aware Attention Mechanism}
Different types of nodes have various impacts on the prediction of adverse drug events.
Inspired by the contextualized entity representation learning from the knowledge supervision of knowledge bases~\citep{yamada2020luke}, we propose to use a concept-aware attention mechanism that distinguishes the types of nodes, especially the concept nodes, and better captures important information related to the positive or negative ADE classes. 

Two types of nodes may not have the same impact on each other. 
Thus, we use different transformations for different types of nodes in the concept-aware attention mechanism in order to learn concept-aware attentive representations. 
We obtain key and value matrices $\mathbf{K}\in \mathbb{R}^{{l}\times{d_h}}$ and $\mathbf{V}\in \mathbb{R}^{{l}\times{d_h}}$ similarly to the key and value in the self-attention of transformer network~\citep{vaswani2017attention}. 
Concept-aware attention has nine different query matrices $\mathbf{Q}$ for concept nodes $c$, word nodes $w$ and document nodes $d$, i.e., $\mathbf{Q}_{ww}$, $\mathbf{Q}_{cc}$, $\mathbf{Q}_{dd}$, $\mathbf{Q}_{cw}$, $\mathbf{Q}_{wc}$, $\mathbf{Q}_{wd}$, $\mathbf{Q}_{dw}$, $\mathbf{Q}_{dc}$, and $\mathbf{Q}_{cd}\in \mathbb{R}^{{l}\times{d_h}}$.
Then, we use $\mathbf{Q}$, $\mathbf{K}$ and $\mathbf{V}$ to compute the attention scores. 
For example, for $i$-th document and $j$-th concept nodes, i.e., $\mathbf{x}_{i}$ and $\mathbf{x}_{j}\in \mathbb{R}^{d_h}$, we calculate the attention score as:
\begin{equation}
	\alpha_{ij}= \operatorname{Softmax} \left( \frac{(\mathbf{K} \mathbf{x}_{j})^{\top} \mathbf{Q}_{cd} \mathbf{x}_{i}}{\sqrt{l}} \right)
\end{equation}
The concept-aware representation $\mathbf{h}_i\in \mathbb{R}^l$ for the $i$-th document is obtained as:
\begin{equation}
	\mathbf{h}_{i}=\sum_{j=1}^{n} \alpha_{i j} \mathbf{V} \mathbf{x}_{j}
\end{equation}
We can obtain the representations of word and concept nodes in the same way. 
These concept-aware representations are fed to the graph network as the node features in the next iteration of model updating.

\subsection{Classification Layers and Model Training}

We apply the two linear layers and a softmax function over the concept-aware document embeddings $\mathbf{h}_i$ to compute the probability of classifying the document in each class  $\mathbf{p}_{g}$, representing the presence or absence of mentions of ADE in the document.
Besides, the interpolation of the prediction probability of two classifiers is adopted to combine the prediction of graph-based modules and pretrained language model-based predictions~\citep{lin2021bertgcn}. 
We use a similar classification module to process the contextualized embeddings from the pretrained language model (the upper branch in Fig. \ref{fig:model_KnowCAGE}) and denote the corresponding classification probabilities by $\mathbf{p}_{c}$.
A weight coefficient $\lambda \in [0,1)$ is introduced to balance the results from graph-based encoding and contextualized models:
\begin{equation}
\mathbf{p} = \lambda \mathbf{p}_{g}+(1-\lambda)\mathbf{p}_{c}.
\end{equation}
This interpolation strategy can also be viewed as a weighted ensemble of two classifiers.

ADE detection is a binary classification task, and the classes are highly imbalanced in most datasets. 
To complicate the matter further, most datasets contain only a small number of samples, making the downsampling method to balance the classes inappropriate. 
This study applies the weighted binary cross-entropy loss function to alleviate this problem. 
The weighted loss function is denoted as:
\begin{equation}
\small
\mathcal{L}=\sum_{i=1}^N \left[-w_{+} y_i\log(p_i)-w_{-}(1-y_i)\log(1-p_i)\right],
\end{equation}
where $w_+= \frac{N_1}{N_0+N_1}$ and $w_-= \frac{N_0}{N_0+N_1}$ are weights of documents predicted as positive or negative samples respectively,  $N_0$ and $N_1$ are the numbers of negative/positive samples in the training set,  and $y_i$ is the ground-truth label of a document. 
The Adam optimizer~\citep{kingma2015adam} is used for model optimization.

\section{Experimental Setup}
\label{sec::experiment}

Our goal is to conduct experiments on four ADE datasets and answer the following research questions. 
\begin{description}
\item[RQ1:] How does the proposed model perform in ADE detection on texts from various sources, compared to other methods? 
\item[RQ2:] How does the heterogeneous graph convolution with knowledge augmentation improve the accuracy of ADE detection?
\item[RQ3:] Does the concept-aware attention improve the accuracy of the heterogeneous graph convolution to detect ADE?
\end{description}
In this section, we will describe the setup of the experiments, and in the next section, we will present the results of the experiments.

\subsection{Data and Preprocessing}
We used four datasets from the medical forum, biomedical publications, and social media, as summarized in Table \ref{tab:datasets}, for evaluation.
We preprocess data by removing stop words, punctuation, and numbers. 
For the data collected from Twitter, we use the tweet-preprocessor Python package~\footnote{\url{https://pypi.org/project/tweet-preprocessor/}} to remove URLs, emojis, and some reserved words for tweets.

\begin{table}[htbp] 
\centering
\small
\caption{A statistical summary of datasets}
\label{tab:datasets}
\begin{tabular}{lcccc}
\toprule
Dataset & Documents & ADE & non-ADE \\
\midrule
SMM4H & 2,418 & 1,209 & 1,209\\
TwiMed-Pub & 1,000 & 191 & 809\\
TwiMed-Twitter & 625 & 232  & 393\\
CADEC & 7,474 & 2,478 & 4,996\\
\bottomrule
\end{tabular}
\end{table}

\noindent \textbf{TwiMed (TwiMed-Twitter and TwiMed-Pub)~\footnote{\url{https://github.com/nestoralvaro/TwiMed}}} The TwiMed dataset~\citep{alvaro2017twimed} includes two sets collected from different domains, i.e., TwiMed-Twitter from social media and TwiMed-Pub for biomedical publications.
In each document, people from various backgrounds annotate diseases, symptoms, drugs, and their relationships. 
A document annotated as outcome-negative is regarded as an adverse drug event.
Models are tested using 10-fold cross-validation.

\noindent \textbf{SMM4H~\footnote{\url{https://healthlanguageprocessing.org/smm4h-2021/task-1/}}} This dataset from Social Media Mining for Health Applications (\#SMM4H) shared tasks~\citep{magge2021overview} is collected from Twitter with a description of drugs and diseases. 
We use the official validation set to evaluate the model performance for a fair comparison with baseline models developed in the SMM4H shared task.

\noindent \textbf{CADEC~\footnote{\url{https://data.csiro.au/collection/csiro:10948}}} The CSIRO Adverse Drug Event Corpus contains patient-reported posts from a medical forum called AskaPatient~\citep{karimi2015cadec}. 
It includes extensive annotations on drugs, side effects, symptoms, and diseases.
We use 10-fold cross-validation to evaluate the model's performance.
\subsection{Baselines and Evaluation}

We compare the performance of our method with two sets of baseline models: 1) models designed for ADE detection and 2) pretrained contextualized models, and report Precision (P), Recall (R), and F1-score.

Customized models for ADE detection are as follows.
\textbf{CNN-Transfer}~\citep{li2020exploiting} (CNN-T for short) and \textbf{ATL}~\citep{li2020exploiting} both exploited adversarial transfer learning, but they applied various feature extractors. 
Variants of attention mechanisms are designed and utilized in \textbf{HTR-MSA}~\citep{wu2018detecting}, \textbf{IAN}~\citep{alimova2018interactive}, \textbf{MSAM}~\citep{zhang2019adverse}, and \textbf{ANNSA}~\citep{zhang2021adversarial}. 
\textbf{CGEM}~\citep{gao2022contextualized} is a predecessor of our work, which developed a contextualized graph-based model.

The previously mentioned ADE detection baselines did not use the SMM4H dataset in their experiments. Therefore, we compare our model with pretrained language models. 
We use the base version of pretrained models for a fair comparison.
\citet{yaseen2021neural} combined the LSTM network with the BERT text encoder~\citep{devlin2018bert} for ADE detection. We denote it as BERT-LSTM.
\citet{pimpalkhute2021iiitn} introduced a data augmentation method and adopted the RoBERTa text encoder with additional classification layers~\citep{liu2019roberta} for ADE detection, denoted as RoBERTa-aug.
\citet{kayastha2021bert} utilized the domain-specific BERTweet~\citep{nguyen2020bertweet} that is pretrained with English Tweets using the same architecture as BERT-base and classified ADE with a single-layer BiLSTM network, denoted as BERTweet-LSTM.

Furthermore, Large Language Models (LLMs) have demonstrated remarkable proficiency across various NLP tasks. 
In this study, we specifically investigate the zero-shot capabilities of \textbf{GPT-4} \citet{openai2023gpt} across four distinct datasets. 
We utilize the following prompt: \textit{"Here are documents from [name and a short description about the dataset]. Determine if the following text contains adverse drug events (ADEs), with the answer being either `yes' or `no'."}.

\subsection{Hyper-parameters}
Table \ref{tab:hyperparameters} shows the hyper-parameters we tuned in our experiments, where LR is the learning rate. 
When the number of iterations exceeds a certain threshold, the learning rate scheduler decays the learning rate by the parameter $\gamma$. 
In our experiment, we set $\gamma$ and the iteration milestone to 0.1 and 30, respectively. 

\begin{table}[ht!]
\centering
\small
\caption{Choices of hyper-parameters}
\label{tab:hyperparameters}
\begin{tabular}{lc}
\toprule
Hyper-parameters & Choices\\
\midrule
LR for text encoder & $2e^{-5}$, $3e^{-5}$, $1e^{-4}$\\
LR for classifier & $1e^{-4}$, $5e^{-4}$, $1e^{-3}$\\
LR for graph-based models & $1e^{-3}$, $3e^{-3}$, $5e^{-3}$\\
Hidden dimension for GNN & 200,~300,~400\\
Weight coefficient $\lambda$ & 0, 0.1 0.3, 0.5, 0.7, 0.9\\
\bottomrule
\end{tabular}
\end{table}

\section{Results}

\subsection{Comparison with Baselines in Different Domains (RQ1)}
\label{sec::result}
We compare our model's predictive performance with baseline models on the TwiMed (Table \ref{tab:TwiMed}), SMM4H (Table \ref{tab:smm4h}) and CADEC (Table \ref{tab:cadec}) datasets.
The results of GPT-4 are obtained in a zero-shot setting, while the results of other baselines are taken directly from the original papers. 
However, some of the baselines did not conduct experiments on all four datasets.
Our proposed model outperforms baseline models, in most cases, demonstrating its effectiveness in ADE detection from texts in various domains (RQ1).
It is worth noting that while GPT-4 shows robust natural language understanding capabilities, its zero-shot performance on this task falls short of the effectiveness demonstrated by our fine-tuned knowledge-augmented model.

We observed that GPT-4 exhibits a tendency to erroneously identify the presence of ADEs in this task, resulting in a notably low recall value. 
Our model achieves a more balanced precision-recall trade-off, leading to higher F1 scores. 
Table \ref{tab:cadec} shows that our model consistently outperforms the baselines. 
Our proposed model can capture rich features to identify a document containing ADEs. 

\begin{table*}[ht!]
\centering
\small
\caption{Results for two TwiMed datasets, i.e., TwiMed-Pub and TwiMed-Twitter. Scores are reported with the mean of 10-fold cross validation following the setup of baselines for our approach. The results of customized models are from the corresponding publications. Results of GPT-4 are derived in a zero-shot setting. \textbf{Bold} text indicates the best performance.}
\label{tab:TwiMed}
\setlength{\tabcolsep}{10pt}
\begin{tabular}{lccc|ccc}
\toprule
\multirow{2}{4em}{Models}   & \multicolumn{3}{c}{TwiMed-Pub} & \multicolumn{3}{c}{TwiMed-Twitter} \\
            & P ($\%$) & R ($\%$) & F1 ($\%$) & P ($\%$) & R ($\%$) & F1 ($\%$)  \\
\midrule
HTR-MSA~\citep{wu2018detecting}  & 75.0 &   66.0            &  70.2 & 60.7 &   61.7            &  61.2       \\    
IAN~\citep{alimova2018interactive}  & 87.8 &   73.8            &  79.2     & 83.6 &   81.3            &  82.4  \\
CNN-T~\citep{li2020exploiting}  & 81.3 &   63.9            &  71.6  & 61.8 &   60.0           &  60.9    \\ 
MSAM~\citep{zhang2019adverse}  & 85.8 &   85.2          &  85.3  & 74.8 &   \textbf{85.6}            &  79.9    \\ 
ATL~\citep{li2020exploiting}  & 81.5 &   67.0            &  73.4   & 63.7 &   63.4           &  63.5     \\ 
CGEM~\citep{gao2022contextualized} & 88.4 & 85.0 & 86.7 & 84.2 & 83.7 & 83.9 \\
GPT-4~\citep{openai2023gpt} & 89.2 & 85.4 & 87.0 & 76.1 & 85.3 & 80.1 \\
\hline
KnowCAGE (GCN) & {88.8} &   \textbf{85.8}            &  \textbf{87.3}  &{84.1} & {84.0} & 84.0 \\
KnowCAGE (GAT) & \textbf{89.6} & {83.4}   & 86.4 & \textbf{84.8} &   84.1            &  \textbf{84.4}   \\
KnowCAGE (DGCNN) & {88.7} & 83.7   & 86.1   &{83.5} & {84.1} & 83.8 \\
\bottomrule
\end{tabular}
\end{table*}

\begin{table}[htbp!]
\centering
\scriptsize
\caption{Results of SMM4H dataset. For our approach, scores are reported for the best-performing results, which follows the setup of baselines. Results from GPT-4 are obtained from a zero-shot setting, while results of other baselines are from the corresponding publications.  \textbf{Bold} text indicates the best performance.}
\label{tab:smm4h}
\setlength{\tabcolsep}{1pt}
\begin{tabular}{lccc}
\toprule
Models   & P ($\%$) & R ($\%$) & F1 ($\%$) \\
\midrule
BERTweet-LSTM~\citep{kayastha2021bert}  & 81.2     & 86.2    & 83.6                \\
RoBERTa-aug~\citep{pimpalkhute2021iiitn}  & 82.1     & 85.7    & 84.3            \\
BERT-LSTM~\citep{yaseen2021neural} & 77.0     & 72.0    & 74.0                \\ 
CGEM~\citep{gao2022contextualized} & 86.7& 93.4 & 89.9 \\
GPT-4~\citep{openai2023gpt} & 62.4& 96.7 & 75.9 \\
\hline
KnowCAGE (GCN) & {85.3} & {95.9} & 90.3  \\
KnowCAGE (GAT) & {85.4} & {94.6} & 89.8  \\
KnowCAGE (DGCNN) &  \textbf{87.2}     & \textbf{97.0}    & \textbf{91.8}                \\
\bottomrule
\end{tabular}
\end{table}

\begin{table}[ht!]
\centering
\scriptsize
\caption{Results for CADEC dataset. For our approach, scores are reported with the mean of 10-fold cross validation following the setup of baselines. Results of GPT-4 are obtained in a zero-shot setting, while results of other baselines are from the corresponding publications. \textbf{Bold} text indicates the best performance.}
\label{tab:cadec}
\begin{tabular}{lccc}
\toprule
Models   & P ($\%$) & R ($\%$) & F1 ($\%$)  \\
\midrule
 HTR-MSA~\citep{wu2018detecting}  & 81.8     & 77.6    & 79.7  \\ 
 CNN-T~\citep{li2020exploiting}  & 84.8     & 79.4    & 82.0  \\
 ATL~\citep{li2020exploiting}  & 84.3     & 81.3    & 82.8  \\
 ANNSA~\citep{zhang2021adversarial}  & 82.7     & 83.5    & 83.1  \\
 GPT-4~\citep{openai2023gpt}  & 68.6     & 83.0    & 75.1  \\
 \hline
 KnowCAGE (GCN) & 86.6 & {90.8}  & 88.7 \\
 KnowCAGE (GAT) & {87.1} & {89.7} & 88.4 \\
 KnowCAGE (DGCNN) & \textbf{87.1}     & \textbf{93.9}    & \textbf{90.4} \\ 
\bottomrule
\end{tabular}
\end{table}

\subsection{Usefulness of the Knowledge Augmented Graph Convolution (RQ2)}
Here, we investigate in more detail how the heterogeneous graph convolution with knowledge augmentation can help with ADE detection (RQ2).
In Table \ref{tab:TwiMed}, most models such as HTR-MSA, IAN, CNN-T and ATL perform worse on TwiMed-Twitter dataset, showing that it is difficult to process informal tweets with colloquial language.  
However, the graph-based encoder in our model helps in effectively encoding information from the informal text, resulting in a better ability to capture the relationships between different entities, improving performance in most cases.
Table \ref{tab:smm4h} compares our model with several pretrained BERT-based models. Our model differs from the pretrained models by addition employing the GNN architectures in addition to the pretrained embeddings, and the results suggest the GNN can further improve models' performance on this task. Compared with another graph-based model, CGEM, our method applies knowledge augmentation to incorporate concept information into graph learning, which is also seen to improve the performance of ADE detection in most cases.

\begin{figure*}[ht!]
\centering
\begin{subfigure}[]{0.24\textwidth}
    \includegraphics[width=\textwidth]{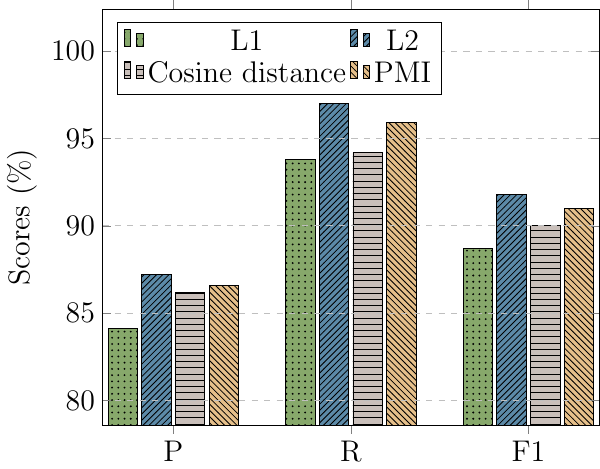}
    \caption{SMM4H}
    \label{fig:wt_smm4h}
\end{subfigure}
\hfil
\begin{subfigure}[]{0.24\textwidth}
    \includegraphics[width=\textwidth]{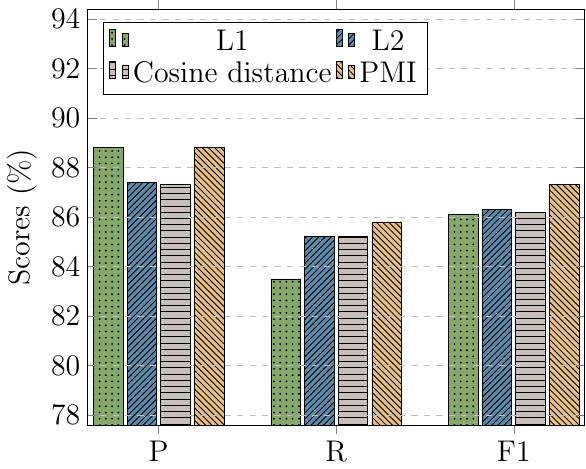}
    \caption{TwiMed-Pub}
    \label{fig:wt_pub}
\end{subfigure}
\hfil
\begin{subfigure}[]{0.24\textwidth}
    \includegraphics[width=\textwidth]{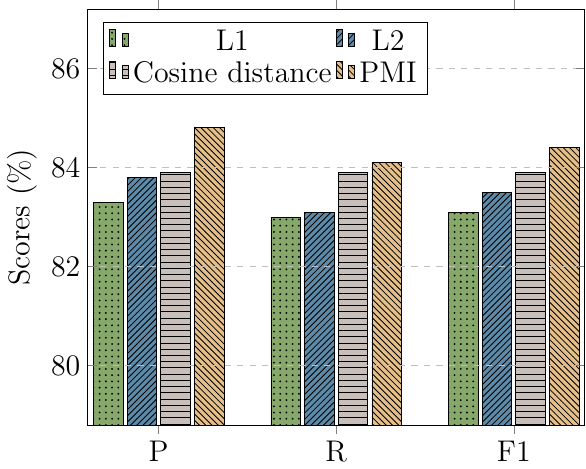}
    \caption{TwiMed-Twitter}
    \label{fig:wt_twi}
\end{subfigure}
\hfil
\begin{subfigure}[]{0.24\textwidth}
    \includegraphics[width=\textwidth]{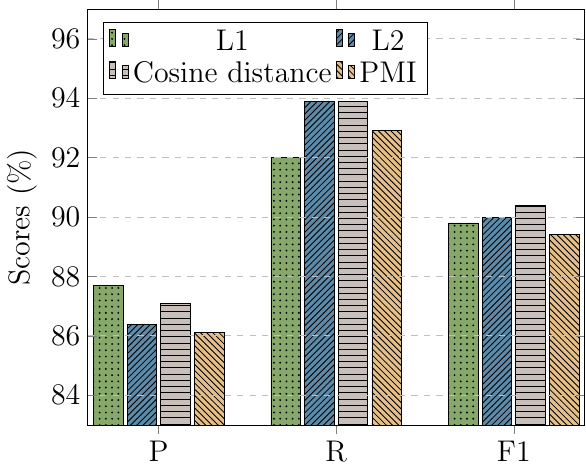}
    \caption{CADEC}
    \label{fig:wt_cadec}
\end{subfigure}
\caption{The effect of different computational methods for weights of edges.}
\label{fig:weights_computation}
\end{figure*}

During the graph construction process, we apply four different computational methods, L1 distance, L2 distance, Cosine distance, and Pointwise Mutual Information (PMI), to derive weights of word-word edges, word-concept edges, and concept-concept edges. 
These methods measure the association between words from various perspectives. 
PMI focuses on the co-occurrence of each word pair, while others consider the words' semantic distances.
Figure \ref{fig:weights_computation} shows the effects of these measurements. 
For TwiMed-Pub and TwiMed-Twitter, the utilization of PMI when graph construction yields the best performance. 
It shows that the incorporation of word co-occurrence information achieves better graph representation for these two datasets.
In contrast, distance-based computational methods of edge weights produce the optimal results for datasets with more training samples, such as SMM4H and CADEC datasets, suggesting that the choice of measurement should be conditional upon the specific characteristics of the datasets.

Additionally, we examine three graph architectures to study which one is most suitable for the ADE detection task. 
Similar to the selection of edge weights, the sample size of the dataset affects the choice of graph architectures. For datasets containing more training samples (i.e., SMM4H and CADEC datasets), DGCNN performs better.
When the number of training samples is small, GCN and GAT achieve better performance.
Hence, we conclude that the graph-based encoding method improves the performance. 
However, we also notice that none of the examined graph architectures consistently outperforms the others on all three datasets from different domains. 
 
\subsection{Effectiveness of the Concept-Aware Attention (RQ3)}
We examine the effectiveness of concept-aware attention by comparing it with two other attention mechanisms, i.e., simple dot-product attention \citep{gao2022contextualized} and structured self-attention \citep{lin2017structured}. 
Table \ref{tab:attention} shows that concept-aware attention consistently achieves the best F1 score on four datasets. 
The concept-aware attention distinguishes different types of nodes from the heterogeneous graph and makes the overall model better utilize the knowledge augmentation from the ULMS.

\begin{table*}[ht!]
\small
\centering
\caption{Comparison on the choices of attention mechanisms}
\label{tab:attention}
\begin{tabular}{lccccccccc}
\toprule
\multirow{2}{8em}{Datasets} & \multicolumn{3}{c}{dot-product attention} & \multicolumn{3}{c}{structured attention} & \multicolumn{3}{c}{concept-aware attention}  \\ 
                  & {P ($\%$)}    & {R ($\%$)}    & F1 ($\%$)   & {P ($\%$)}    & {R ($\%$)}    & F1 ($\%$)   & {P ($\%$)}    & {R ($\%$)}    & F1 ($\%$)    \\ 
\midrule
SMM4H           & {86.0} & {94.2} & 89.9 & {85.0} & {95.4} & 89.3 & {89.9} & {97.0} & \textbf{91.8}  \\
TwiMed-Pub     & {87.9} & {84.5} & 86.2 & {88.9} & {82.9}   & 85.8 & {88.8} & {85.8} & \textbf{87.3} \\
TwiMed-Twitter          & {84.5} & {82.2} & 83.4 & 83.0 & 81.8   &82.4   & {84.8} & {84.1} & \textbf{84.4}  \\
CADEC           & {86.7} & {89.1} & 87.9 & {84.3} & {88.2}   & 86.2   & {87.1} & {93.9} & \textbf{90.4}  \\
\bottomrule
\end{tabular}
\end{table*}

\subsection{Case Study}
By adopting Concept-Aware attention, crucial words and concepts can be effectively highlighted, which ultimately improves the classification performance of the model. 
Meanwhile, this attention mechanism also improves the model's explainability, as it provides explicit insight into the importance of words and concepts through their corresponding attention scores.  
In this section, we present a case study to demonstrate the effect of Concept-Aware Attention. 
We select one positive sample and one negative sample from the SMM4H dataset, respectively. 
They are both accurately classified by the proposed model, where Concept-Aware Attention allows important words and concepts to be highlighted.

\begin{figure}[ht!]
\centering
\begin{subfigure}{0.34\textwidth}
    \includegraphics[width=\textwidth]{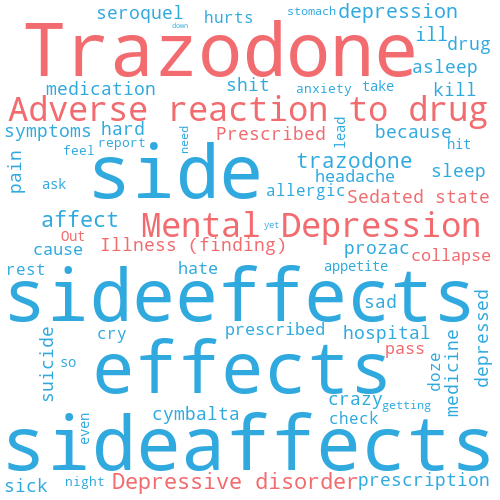}
    \caption{An example of a positive document}
    \label{fig:wc_ade}
\end{subfigure}
\hfil
\begin{subfigure}{0.34\textwidth}
    \includegraphics[width=\textwidth]{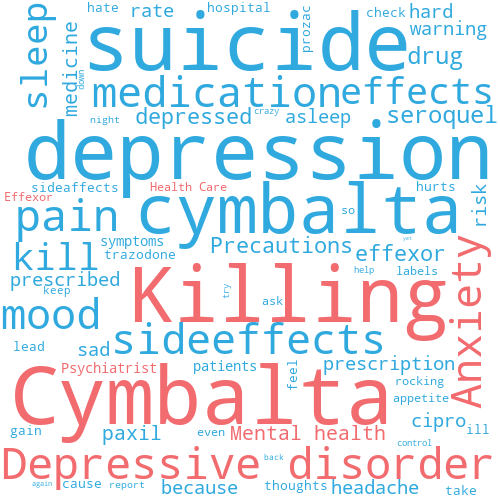}
    \caption{An example of a negative document}
    \label{fig:wc_nonade}
\end{subfigure}
\caption{A visualization of the concept-aware attention in the form of node cloud }
\label{fig:wc}
\end{figure}

We visualize it by employing the node cloud. In Figure \ref{fig:wc}, distinct colors distinguish word nodes from concept nodes. 
Node sizes correspond to their contribution when classifying the document; thus, nodes that are more important to the document will be shown in larger sizes. 
We set a threshold to ensure that only nodes of sufficient importance will be in the figure.

Figure \ref{fig:wc_ade} presents the node cloud for the document ``\textbf{debating on taking a trazodone and literally passing out for the day}''. 
The document is a positive sample containing the implications of the adverse effects.
The figure's blue-shaded word nodes can be grouped into five categories: words containing the meaning of ``side effects''; names of medicines for depression, adverse effects; depression symptoms; and others. 
The model learns enough knowledge related to the symptoms and side effects of a certain disease. 
Moreover, the integration of concept-aware attention enhances the model's focus on concepts indicating adverse effects. 
Red concept nodes shown in the figure are identified to be closely related to the document by the model. They are centered around side effects and depression, emphasizing that the incorporation of concept-aware attention facilitates a better understanding of concept-level information by the model.

The node cloud for the negative document ``\textbf{i don't get why they make stuff like cymbalta to decrease the suicide rate, when there are bunch of precautions about how it may kill you.}'' is visualized in Figure \ref{fig:wc_nonade}. 
Negative documents are more likely to be misclassified in this task than positive documents. 
Because negative documents often contain: (a) descriptions of disease symptoms, which are easily mistaken for adverse drug effects; (b) negative sentiments towards drugs, without descriptions of adverse effects.
The example in Figure \ref{fig:wc_nonade} corresponds to case (b). 
However, with the incorporation of concept-aware attention, the model focuses more on the word nodes and concept nodes that represent the meaning of the document itself without mistakenly associating information related to adverse effects.
Therefore, contrary to the example in Figure \ref{fig:wc_ade}, most nodes shown in Figure \ref{fig:wc_nonade} lack a direct relation to ``adverse effects''. 
Instead, these nodes mainly correspond to symptoms of depression and the term ``suicide''.

\section{Conclusion}
\label{sec::conclusion}
The automated detection of adverse drug events from social media content or biomedical literature requires the model to encode text information and capture the relation between drugs and adverse effects efficiently. 
This paper utilizes knowledge-augmented contextualized graph embeddings to learn contextual information and capture relations for ADE detection. 
We equip different graph convolutional networks with pretrained language representations over the knowledge-augmented heterogeneous text graph and develop concept-aware attention to optimally process the different types of nodes in the graph.
By comparing our model with other baseline methods, experimental results show that graph-based embeddings incorporating concept information from the UMLS can inject medical knowledge into the model and the concept-aware attention can learn richer concept-aware representations, leading to better detection performance.

\section{Acknowledgements}
We acknowledge the computational resources provided by the Aalto Science-IT project and CSC - IT Center for Science, Finland.
This work was supported by the Research Council of Finland (Flagship programme: Finnish Center for Artificial Intelligence FCAI, and grants 336033, 352986, 358246) and EU (H2020 grant 101016775 and NextGenerationEU).
We thank Volker Tresp, Zhen Han, Ruotong Liao, and Zhiliang Wu for their valuable discussions. 

\section{Bibliographical References}\label{sec:reference}

\bibliographystyle{lrec-coling2024-natbib}
\bibliography{ADE-detection}

\end{document}